\RequirePackage[OT1]{fontenc}
\documentclass[conference]{IEEEtran}

\usepackage{cite} 
\usepackage[hidelinks,pdfusetitle]{hyperref}
\usepackage[acronym]{glossaries}
\usepackage[inline]{enumitem}
\usepackage{amsmath,amssymb,amsfonts}
\usepackage{graphicx}
\usepackage{layouts}
\usepackage{multirow}
\usepackage{shellesc}
\usepackage{subcaption}
\usepackage{textcomp}
\usepackage{xcolor}
\usepackage{tikzscale}
\usepackage{tikz}
\usepackage{flushend}

\usetikzlibrary{external}
\makeatletter
\hypersetup{
	pdfauthor={J{\o}rgen Nordmoen, Frank Veenstra, Kai Olav Ellefsen and Kyrre Glette},
	pdfsubject={arXiv submission},
	pdfkeywords={modular robotics, quality diversity, map-elites, NSGA-II, comparison}
}
\makeatother

\begin{document}
\newacronym{cpg}{CPG}{Central Pattern Generator}
\newacronym{ea}{EA}{Evolutionary Algorithm}
\newacronym{er}{ER}{Evolutionary Robotics}
\newacronym{map-elites}{MAP-Elites}{Multi-dimensional Archive of Phenotypic Elites}
\newacronym{moea}{MOEA}{Multi-Objective Evolutionary Algorithm}
\newacronym{nsga}{NSGA-II}{Non-dominated Sorting Genetic Algorithm-II}
\newacronym{qd}{QD}{Quality Diversity}

\title{Quality and Diversity in Evolutionary Modular Robotics}

\makeatletter
\newcommand{\linebreakand}{%
  \end{@IEEEauthorhalign}
  \hfill\mbox{}\par
  \mbox{}\hfill\begin{@IEEEauthorhalign}
}
\makeatother

\author{
\IEEEauthorblockN{1\textsuperscript{st} J{\o}rgen Nordmoen}
\IEEEauthorblockA{\textit{Department of Informatics} \\
\textit{University of Oslo}\\
Oslo, Norway\\
jorgehn@ifi.uio.no}
\and
\IEEEauthorblockN{2\textsuperscript{nd} Frank Veenstra}
\IEEEauthorblockA{\textit{Department of Informatics \& RITMO} \\
\textit{University of Oslo}\\
Oslo, Norway}
\and
\IEEEauthorblockN{3\textsuperscript{rd} Kai Olav Ellefsen}
\IEEEauthorblockA{\textit{Department of Informatics} \\
\textit{University of Oslo}\\
Oslo, Norway}
\linebreakand
\IEEEauthorblockN{4\textsuperscript{th} Kyrre Glette}
\IEEEauthorblockA{\textit{Department of Informatics \& RITMO} \\
\textit{University of Oslo}\\
Oslo, Norway}
}

\maketitle
\begin{abstract}
	In \acrlong{er} a population of solutions is evolved to optimize robots
	that solve a given task. However, in traditional \acrlong{ea}s, the
	population of solutions tends to converge to local optima when the
	problem is complex or the search space is large, a problem known as
	premature convergence. \Acrlong{qd} algorithms try to overcome premature
	convergence by introducing additional measures that reward solutions for
	being different while not necessarily performing better.
	In this paper we compare a single objective \acrlong{ea} with two
	diversity promoting search algorithms; a \acrlong{moea} and
	\acrshort{map-elites} a \acrlong{qd} algorithm, for the difficult
	problem of evolving control and morphology in modular robotics. We
	compare their ability to produce high performing solutions, in addition
	to analyze the evolved morphological diversity.
	The results show that all three search algorithms are capable of
	evolving high performing individuals. However, the \acrlong{qd}
	algorithm is better adept at filling all niches with high-performing
	solutions. This confirms that \acrlong{qd} algorithms are well suited
	for evolving modular robots and can be an important means of generating
	repertoires of high performing solutions that can be exploited both at
	design- and runtime.
\end{abstract}


\section{Introduction}
For many real-world robotics problems knowing the correct design of the robot's
body and control system in advance can be a difficult challenge. Ideally, we
would like the robot to adapt itself to the task, which in many environments
could also necessitate a change in the robot's morphology. Modular robots are a
class of robots that are comprised of several modules, which in total make up
the morphology of the robot. Such robots, in addition to advances within
3D-printing technology, such as better materials, higher speeds, and increased
portability, could be capable of repairing and/or producing new parts in situ to
adapt to different problems~\cite{revzen2011structure}.

\begin{figure}
	\centering
	\includegraphics{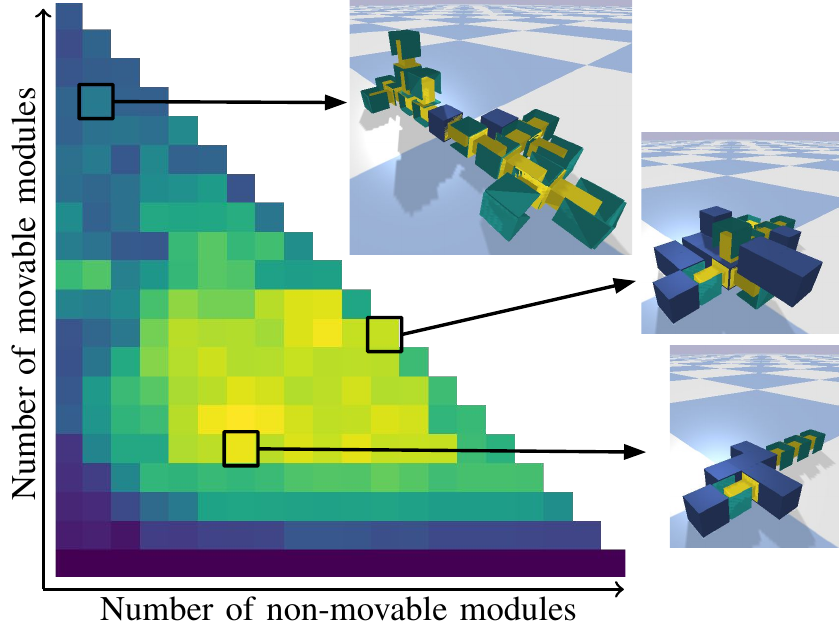}
	\caption{Evolving a varied set of high performing modular robots
	utilizing the \acrlong{qd} paradigm.}
	\label{fig:illustration}
\end{figure}

\Gls{er} tries to solve the problem of automatic design and optimization through
the use of \glspl{ea} which are population-based search algorithms inspired by
natural evolution~\cite{floreano2008bio}. While \gls{er} is not solely focused
on optimizing morphology and control, several prominent examples have shown that
it is not only possible but also that the algorithms employed can exhibit a
surprising amount of
variation~\cite{lipson2000automatic,zykov2007evolved,yim2007modular}.

Several challenges exist when evolving morphology and control for modular
robots. One challenge is how to encode the robot morphology when the topology is
not fixed~\cite{samuelsen2014some}. Another is the choice of control
architecture which can range from simple open-loop wave
generators~\cite{veenstra2017evolution}, to interconnected pattern producing
generators~\cite{marbach2005online} and even more complex neural
networks~\cite{lipson2000automatic}. A third challenge is the problem of
premature convergence, which is a challenge for all classes of \glspl{ea} but
is especially prominent in modular \gls{er} due to the large and complex search
space resulting from the difficult and interconnected relationship of optimizing
morphology and control simultaneously~\cite{lipson2000automatic}.

Overcoming premature convergence in modular \gls{er} require search algorithms
that are not only capable of optimization, but also exploration. In other words,
the search algorithm needs to exhibit both quality and diversity. Different
approaches to this challenge exist~\cite{mouret2012encouraging}, such as
utilizing the notion of heterogeneity to retain solutions based on attributes
other than fitness~\cite{mouret2009overcoming}. \Gls{qd} algorithms are a
sub-set of these diversity promoting algorithms where explicit phenotypic
feature descriptors are used to distinguish and group
solutions~\cite{pugh2016quality}. \Gls{qd} algorithms have been shown to
increase performance and maintain a diverse population when effective feature
descriptors can be defined~\cite{tarapore2016different}.
 
In this paper we compare the quality of evolved solutions of two diversity
promoting search algorithms, the \gls{moea} \gls{nsga}~\cite{deb2002fast} with
diversity as additional objectives and
\gls{map-elites}~\cite{mouret2015illuminating}, with a traditional single
objective \gls{ea} for the difficult task of evolving control and morphology in
modular \gls{er}. Our goal is to understand which attributes of the different
search algorithms contribute to the performance of the systems and to which
capacity \gls{qd} algorithms can be applied to modular \gls{er} challenges.
Because of the explicit definition of feature descriptors utilized in
\gls{map-elites} we believe this algorithm is well suited for modular \gls{er}
as it should be capable of evolving a repertoire of diverse high performing
morphologies. The repertoire should aid in overcoming premature convergence in
addition to serving as a store that can be utilized at design or runtime to
select the best morphology for the given
task~\cite{cully2015robots,gaier2017data}. To group solutions, we utilize the
number of movable and the number of non-movable modules within the morphology as
illustrated in Figure~\ref{fig:illustration}. The selection of morphological
feature descriptors to distinguish solutions, both in \gls{map-elites} and
\gls{nsga}, is a difficult challenge with many
possibilities~\cite{miras2018search}. For the experiments carried out in this
paper, we kept the descriptors simple to avoid confounding factors. By keeping
the feature descriptors simple we focus on the search algorithm's capacity for
evolving diverse solutions and not on the feature descriptors
themselves~\cite{samuelsen2014some}.

Our results show that the different search algorithms are capable of achieving
the same performance for this modular \gls{er} locomotion task. However, the
results also show that the addition of diversity measures in both \gls{nsga} and
\gls{map-elites} greatly increases morphological diversity. The \gls{qd}
algorithm is additionally able to evolve a full repertoire of high performing
solutions showing that \gls{qd} algorithms are appropriate for modular robotics
tasks.

The contribution of our paper is a comparison between a single objective
\gls{ea} and two diversity promoting \glspl{ea} for the difficult problem of
optimizing morphology and control in modular \gls{er}. In the comparison, we
elucidate both the differences between the search algorithms in addition to
comparing the differences in evolved populations.

\section{Background}
In this section relevant background information about modular \gls{er} and
\gls{qd} algorithms will be described.

\subsection{Modular Robotics}
Evolving body and control for artificial creatures have a long history in the
field of Artificial Life~\cite{sims1994evolving}. Modular robotics is
distinguished from these virtual creatures by comprising the morphology of
re-usable homogeneous or heterogeneous building blocks, called
modules~\cite{moubarak2012modular,veenstra2017evolution}. Modular robots
provide a way to effectively transition from simulation to reality since modules
can be fabricated individually and then combined based on designs optimized in
simulation~\cite{stoy2006deformatron}. 

One of the challenges with modular robotics is the interconnected relationship
between control and morphology~\cite{lipson2000automatic, lipson2016difficulty}.
To overcome this challenge many different approaches such as generative
encoding~\cite{hornby2003generative} and different control
architectures~\cite{haasdijk2010hyperneat, marbach2005online} have been applied.

\subsection{\acrlong{qd}}
\Gls{qd} algorithms emerged from the realization that optimization through
diversity can yield high performing solutions and, more importantly, can be
better suited to exploring the whole problem space~\cite{lehman2008exploiting}.
By focusing on phenotypic diversity, \gls{qd} algorithms search the space of
possible solutions without constraining the search to only finding better-fit
solutions~\cite{pugh2016quality}. This separates \gls{qd} algorithms from
traditional \glspl{moea} since Pareto dominated solutions can be kept as long as
their phenotypic expression is sufficiently different from other solutions in
the population~\cite{mouret2015illuminating}.

An interesting property of \gls{qd} algorithms is the capability to produce a
repertoire of different solutions for the same problem~\cite{cully2017quality}.
The repertoire can be exploited, either at design time~\cite{gaier2017data} or
during operation~\cite{cully2015robots}, to select different solutions
depending on the circumstances of the situation.

Although \gls{qd} algorithms have been applied to the evolution of artificial
creatures~\cite{lehman2011evolving} and morphological descriptors have been
used to evolve modular robots~\cite{samuelsen2014some, samuelsen2015real} few
examples exist applying the \gls{qd} paradigm to modular robotics. This makes
our contribution valuable, opening up a new application area for \gls{qd} and
introducing a way to generate a repertoire of possible morphologies within
modular robotics that can later be experimented on in the real world.

\section{Methods}
To compare the different search algorithms, experiments were carried out to
evolve both the control and morphology of a modular \gls{er} system. The main
objective of the search algorithms is to evolve modular robots that locomote
across a plane as quickly as possible. To measure quality in the algorithms,
solutions are compared based on fitness. To ensure that the comparison is as
fair as possible the parameters for the individual algorithms were optimized in
advance and the number of fitness evaluations is kept constant between the
different search algorithms. The parameter search consisted of testing all
permutations of combinations from table~\ref{tab:parameter_search} for all three
search algorithms, giving a total of $241$ combinations to test.

\begin{table}
    \centering
    \begin{tabular}{|c|r|}
	    \hline
	    Parameter & Values \\
	    \hline\hline
	    Probability of morphological mutation & $[0.05, 0.1, 0.2]$ \\
	    Crossover probability & $[0.05, 0.1, 0.2]$ \\
	    Probability of controller mutation & $[0.05, 0.1, 0.2]$ \\
	    Controller standard deviation & $[0.01, 0.05, 0.1]$ \\
	    \hline
    \end{tabular}
    \caption{Description of parameters optimized through a pre-experiment
	parameter search.}
    \label{tab:parameter_search}
\end{table}

In the next section, the modular robots with their morphological encoding and
control system will be described. Following that, the \glspl{ea} will be
described along with the parameters and experiment
configurations. 

\subsection{Modular ER System}\label{sec:modular}
The morphological encoding employed is a tree-based direct encoding similar
to~\cite{faina2013edhmor}. The encoding allows for any combination of modules
representable as a graph where each node in the graph is a module and each edge
is a connection between two modules. For the experiments carried out in this
paper two different modules were utilized, one non-movable rectangular module
supporting $5$ child modules and one servo module capable of moving one side
back-and-forth and supporting $3$ child modules~\cite{moreno2017emerge}, see
Figure~\ref{fig:illustration} for a small selection of robots. Each morphology
starts with a single rectangular module as its root. To randomly initialize the
morphology, a random size is selected between $1$ and $\eta$ (see
Table~\ref{tab:morph_params}). Modules are then added to the tree at random
locations until the size of the morphology equals the selected size.

The morphological encoding supports mutation- and crossover-operators. When
mutating the morphology three possibilities exist
\begin{enumerate*}
    \item \textit{Add a random module.} The tree is traversed and each available
	    connection point is added as a possibility. A
		connection point is randomly selected along with a randomly
		selected module type before being inserted into the tree.
    \item \textit{Remove a module.} The tree is traversed adding all modules
	    except the root into a list of candidates to remove. A module is
		randomly selected from the candidates before being removed along
		with any existing children.
    \item \textit{Mutate a module.} The two modules in use both support rotation
	    around its connection axis and mutation will randomly select a new
		orientation in $90^o$ increments.
\end{enumerate*}
Note that only one of the three possibilities can happen per morphological
mutation.

For crossover, a branch exchange is implemented. For both parent morphologies
the tree is traversed adding all modules, except the root, to a list of
candidates. A random candidate is selected from both morphologies before being
exchanged. The candidate module, including its children, from the first
morphology, is inserted into the place of the candidate from the second
morphology and vice versa.

Lastly, the morphology is limited to a maximum size, $\eta$ (see
Table~\ref{tab:morph_params}), and a maximum depth $\delta$, so that additional
modules are not realized in the simulator. This limit ensures that morphologies
do not grow unbounded and are feasible to simulate.

\begin{table}
    \centering
    \begin{tabular}{|c|c|r|}
        \hline
       Parameter & Description & Value \\
	    \hline\hline
       $\eta$ & Maximum module count & $20$ \\
       $\delta$ & Maximum module depth & $4$ \\
       \hline
    \end{tabular}
    \caption{Morphology parameters.}
    \label{tab:morph_params}
\end{table}

The control system of the modular robots is based on a decentralized wave
pattern controller~\cite{veenstra2017evolution}. Each movable module in the
morphology is given a controller which outputs the desired angle of the joint,
$\theta_i$, according to the following equation

\begin{equation}
    \theta_i = \alpha_i * \sin\left(\omega_i t + \phi\right) + o_i
\end{equation}

where $\alpha_i$ is the amplitude, $t$ is the time since the controller was
initialized, $\omega_i$ is the frequency, $\phi_i$ is the phase offset and $o_i$ is
the amplitude offset for joint module $i$. The output, $\theta_i$, is further
limited to the minimum and maximum angle of the joint so that the control values
do not exceed allowable set-points for the real-world equivalent. The
parameters, and allowable range for each, are summarized in
Table~\ref{tab:ctrl_parameters}.

\begin{table}
    \centering
    \begin{tabular}{|c|c|r|}
        \hline
       Parameter & Description & Range \\
	    \hline\hline
       $\theta$ & Set-point angle & $[-1.57, 1.57]$ \\
       $\alpha$ & Amplitude & $[-1.57, 1.57]$ \\
       $\omega$ & Frequency & $[0.2, 2]$ \\
       $\phi$ & Phase offset & $[-2\pi, 2\pi]$ \\
       $o$ & Offset & $[-1.57, 1.57]$ \\
       \hline
    \end{tabular}
    \caption{Decentralized wave pattern controller parameters. The ranges are
	based on the real world servo used in the modules.}
    \label{tab:ctrl_parameters}
\end{table}

Controllers are mutated separately from morphology and each parameter is
perturbed using Gaussian noise, $\mathcal{N}(p, \sigma)$ where $p$ is the mean and
$\sigma$ is the magnitude of the noise. The magnitude, $\sigma$, is specified
for each search algorithm as a number in $[0, 1]$ and then scaled to the range
of each parameter shown in Table~\ref{tab:ctrl_parameters}. To avoid mutating
parameters outside the allowable range, a bounce-back function is applied
according to the following equation

\begin{equation}
    L(v, min, max) =
    \begin{cases}
    min + (min - v) \text{ if } v < min\\
    max - (v - max) \text{ if } v > max\\
    v \text{ otherwise}
    \end{cases}
\end{equation}

where $v$ is the parameter to mutate, $min$ and $max$ are the parameter's range
taken from Table~\ref{tab:ctrl_parameters}. The effect of this function is to
limit the parameters to their allowable range with a uniform
distribution~\cite{nordmoen2020restricting}.

The fitness function used is based on the straight-line distance between the
starting- and final position of the root module. To discourage optimization
towards local optima, where the robot simply falls over, the selection of
starting point is delayed in time so that early movement is discounted towards
the total fitness. The distance calculation is only performed in the $X$ and $Y$
axis since we are mostly interested in distance on the surface plane. The
evaluation parameters can be found in Table~\ref{tab:parameters}.

The feature descriptors used for \gls{nsga} and \gls{map-elites} are given as
the tuple

\begin{equation}\label{eq:feature_descriptors}
	b' = (m_i, j_i)
\end{equation}

using the notation presented in \cite{mouret2015illuminating}, where $m_i$ is
the number of non-movable modules and $j_i$ is the number of movable joint
modules.

\subsection{Evolutionary Algorithms}\label{sec:evolutionary}
As a baseline a single objective \gls{ea} is selected. The \gls{ea} is
$(\lambda,\mu)$ generational replacement strategy
from~\cite{eiben2003introduction} with tournament selection between two
individuals. The \gls{ea} has a single objective function that is set to the
fitness function described in the previous section. Further configuration
parameters can be found in Table~\ref{tab:parameters}. Note that the population
size is selected to be similar to the maximum number of solutions in the
\gls{map-elites} repertoire.

The first diversity promoting search algorithm is the \gls{moea} \textemdash{}
\gls{nsga}~\cite{deb2002fast}. This search algorithm is used so that diversity
metrics can be introduced into a purely optimizing
\gls{moea}~\cite{mouret2009overcoming}. The diversity metrics are based on the
same morphological descriptors as \gls{map-elites} uses, but for \gls{nsga} to
optimize these objectives they are recast as the sum of difference in diversity
between individuals in the population, \cite{lehman2011evolving}, according to
the following equations

\begin{equation}\label{eq:distance}
    D(x) = \frac{1}{|P_n|}\sum_{y \in P_n} d(x, y)
\end{equation}

\begin{equation}\label{eq:distance1}
    d(x, y) = 1.0 - e^{-|(m_x, j_x) - (m_y, j_y)|}
\end{equation}

where $P_n$ is the population, $x$ and $y$ are solutions in $P_n$, $m_i$ is
the number of non-movable modules and $j_i$ is the number of movable joint
modules in the respective solutions.  The output of
equation~\ref{eq:distance1} is in $\mathbb{R}^2$ giving a total of three
dimensions to optimize with \gls{nsga}. Note that the diversity score is
re-calculated every time a change in the population occurs. It is also important
to point out that equation~\ref{eq:distance1} is altered compared to previous
work~\cite{lehman2011evolving,samuelsen2014some} as we experienced that the
original equation leads to convergence in morphologies\footnote{The convergence
is most likely a result of the maximization of diversity, which leads to
convergence at the morphological extremities.}. The changes to the distance
function, equation~\ref{eq:distance1}, weigh all changes to morphology equally
which mitigates this convergence.

The last search algorithm used is \gls{map-elites}. This algorithm represents
the \gls{qd} paradigm and differs from traditional \glspl{moea} in that the
additional feature descriptors are not optimized, but rather differentiate
solutions for storage in a repertoire. Central to the \gls{map-elites} algorithm
is the notion of feature descriptors which are used as additional objectives for
the search, however, these are not maximized nor minimized. As
described before, we utilize morphological metrics as feature descriptors. We
define the repertoire to contain individuals using the feature descriptors
defined in equation~\ref{eq:feature_descriptors}. The range is set to the
maximum number of modules, described in Table~\ref{tab:morph_params}, so that
the repertoire potentially can contain any morphology representable within those
limits, where the minimal morphology is simply the root module.

\begin{table}[t]
    \centering
    \begin{tabular}{|l|l|r|}
        \hline
	Parameter & Applied to & Value \\ \hline\hline
        Evaluation time & \multirow{5}{*}{Shared} & $20$ seconds \\
        Warm-up before start & & $2$ seconds \\
        Repetitions & & $30$ \\
        Number of generations & & $500$ \\
        Number of evaluations & & $100\ 000$ \\
        \hline
        \multirow{3}{*}{Initial population size} & \acrshort{ea} & \multirow{2}{*}{200} \\
        & \acrshort{nsga} & \\
        & \acrshort{map-elites} & 1000 \\
        \hline
        \multirow{3}{*}{Population / Batch size} & \acrshort{ea} & \multirow{3}{*}{200} \\
        & \acrshort{nsga} & \\
        & \acrshort{map-elites} & \\
        \hline
        \multirow{3}{*}{Selection} & \acrshort{ea} & \multirow{2}{*}{Tournament} \\
        & \acrshort{nsga} & \\
        & \acrshort{map-elites} & Uniform \\
        \hline
        \multirow{3}{*}{Morphological mutation} & \acrshort{ea} & $0.2$ \\
        & \acrshort{nsga} & $0.05$ \\
        & \acrshort{map-elites} & $0.2$ \\
        \hline
        \multirow{3}{*}{Crossover rate} & \acrshort{ea} & $0.2$ \\
        & \acrshort{nsga} & $0.1$ \\
        & \acrshort{map-elites} & $0.2$ \\
        \hline
        \multirow{3}{*}{Controller mutation} & \acrshort{ea} & $0.2$ \\
        & \acrshort{nsga} & $0.2$ \\
        & \acrshort{map-elites} & $0.1$ \\
        \hline
        \multirow{3}{*}{Controller $\sigma$} & \acrshort{ea} &
	    $0.05$ \\
        & \acrshort{nsga} & $0.1$ \\
        & \acrshort{map-elites} & $0.1$ \\
        \hline
    \end{tabular}
    \caption{Experiment parameters.}
    \label{tab:parameters}
\end{table}

\section{Results}
To understand the quality of the three search algorithms the maximum
fitness in each repetition was recorded. In Figure~\ref{fig:fitness} the
trajectory of each algorithm is shown over evolutionary time and
Figure~\ref{fig:max_fitness} shows the full distribution of the last generation.
From the fitness gradients it can be seen that the single objective \gls{ea}
more quickly finds fit solutions, while the two diversity promoting search
algorithms take more time. We can also see that \gls{map-elites} has less
variation across the different runs of the experiments.

When comparing the full distribution, shown in Figure~\ref{fig:max_fitness},
through a pairwise Wilcoxon rank sum test with Holm~\cite{holm1979simple}
correction, statistical significant differences can be found between the single
objective \gls{ea} and \gls{nsga} and \gls{map-elites} and \gls{nsga}.
Furthermore a Fligner-Killeen test of homogeneity of
variances~\cite{conover1981comparative} shows statistical significant
differences between the three algorithms. In other words, there is an observable
difference in the mean and distribution of the three search algorithms.

\begin{figure*}
    \begin{subfigure}{0.68\textwidth}
        \caption{}
	\includegraphics{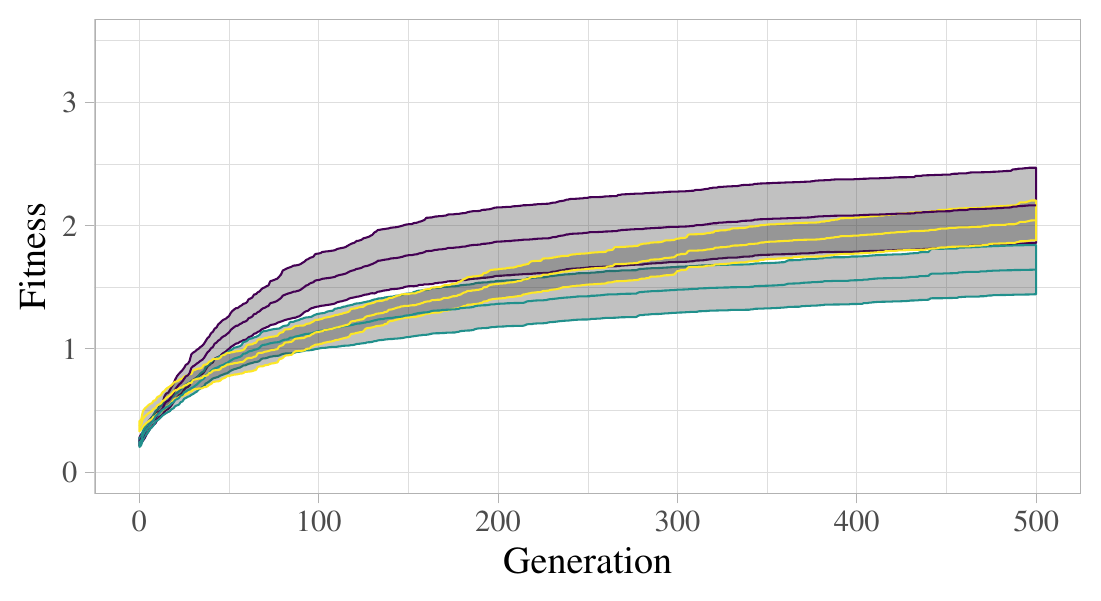}
        \label{fig:fitness}
    \end{subfigure}
    \hspace*{-30pt}
    \begin{subfigure}{0.3\textwidth}
        \caption{}
	\includegraphics{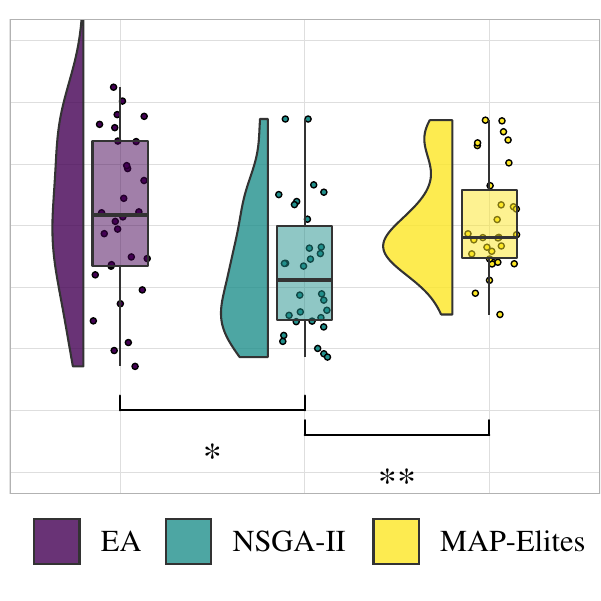}
        \label{fig:max_fitness}
    \end{subfigure}
	\vspace*{-20pt}
	\caption{Maximum fitness discovered in each run. In (a) the mean is
	shown with a $95\%$ confidence interval while (b) shows the full
	distribution after the last evaluations. Statistical significant
	differences are noted in (b) using a Pairwise Wilcoxon Rank Sum test
	with Holm~\cite{holm1979simple} correction.}
    \label{fig:fitness_all}
\end{figure*}

Since we are interested in finding a diverse set of high-performing solutions
and two of the search algorithms are able to utilize morphological descriptors
to encourage diversity it is instructive to project the results along these
dimensions. To project the population into the repertoires, as used in
\gls{map-elites}, we create an empty repertoire at the start of evolution and
insert solutions as they appear. For the single objective \gls{ea} this means
that solutions can be retained in the repertoire longer than it was kept in the
population as long as the fitness is better than newer solutions with the same
feature descriptors. Figure~\ref{fig:map_evolution} shows a selection of these
projections through time. The first row shows a single run, selected as the run
closest to the median best in Figure~\ref{fig:max_fitness}, exemplifying the
output one could expect when running each search algorithm once. The next row
shows a heatmap of the number of runs that found a solution for each
morphological description. Finally the last row shows the average fitness of
each morphological description where the solution is the cumulative best found
in each run of the experiment. Note that the color gradient used changes scale.
As can be seen from the heatmaps, \gls{map-elites} is able to fill out every
niche in almost all runs. From the last row it can also be seen that
\gls{map-elites} is effective, on average, at finding high performing solutions
compared to the other search algorithms. When comparing \gls{nsga} and the
single objective \gls{ea}, we can see the effect of adding diversity measures to
an optimizing \gls{ea}, where \gls{nsga} is able to find a more diverse set of
solutions.

\begin{figure}[t]
    \begin{subfigure}{0.45\textwidth}
	\includegraphics{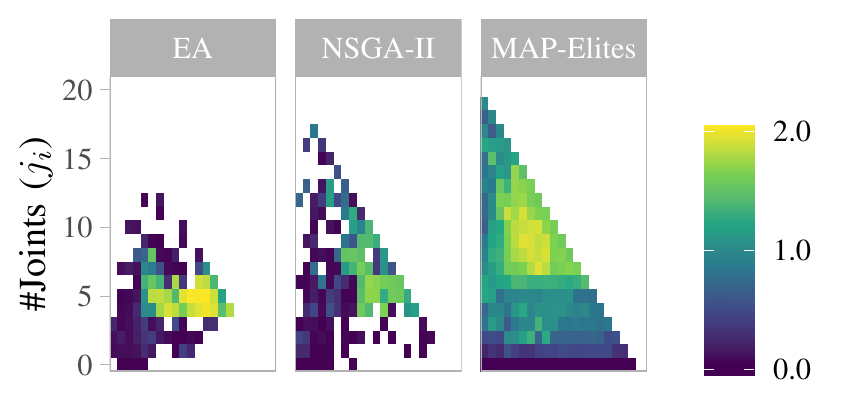}
        \label{fig:map0}
	\vspace*{-18pt}
    \end{subfigure}
    \begin{subfigure}{0.45\textwidth}
	\includegraphics{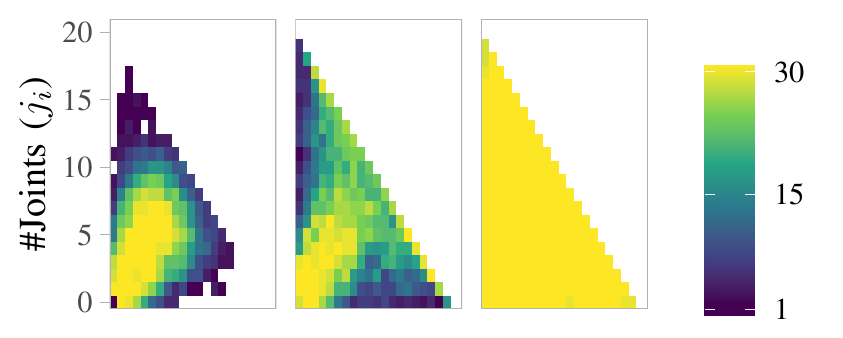}
        \label{fig:map250}
	\vspace*{-18pt}
    \end{subfigure}
    \begin{subfigure}{0.45\textwidth}
	\includegraphics{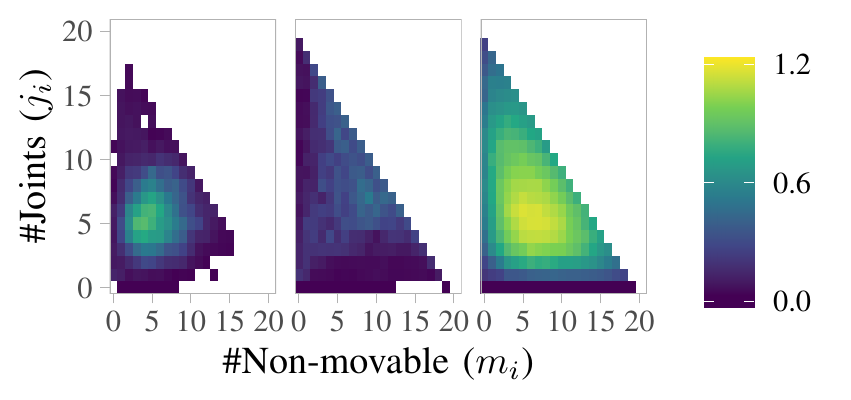}
        \label{fig:map500}
    \end{subfigure}
	\vspace*{-20pt}
    \caption{Projection of the solutions found for each morphological
	description throughout the experiment. The top row shows a single
	representative run to give an impression of the expected outcome. The
	next row shows the number of runs that found a solution for the given
	morphological description. The last row shows the mean fitness of the
	best individuals over all runs.}
    \label{fig:map_evolution}
\end{figure}

The projections shown in Figure~\ref{fig:map_evolution} can further be
summarized through methods developed within the \gls{qd}
paradigm~\cite{pugh2016quality}. The \gls{qd}-score calculation, shown in
Figure~\ref{fig:qd_score}, summarizes the total fitness of all solutions in the
map projection according to the following equation

\begin{equation}
	\text{QD - score}(m) = \sum_{x \in m} Q_x
\end{equation}

where $m$ is a projection, $x$ is a solution in the projection and $Q_x$ is the
quality of solution $x$. The metric gives a good balance between exploring the
search space and exploiting already found solutions and can be better at
comparing \gls{qd} algorithms than earlier \textit{precision} and
\textit{coverage} plots. Figure~\ref{fig:qd_score} demonstrates that
\gls{map-elites} is able to evolve a more diverse set of high-performing
solutions compared to the other two search algorithms. For the single objective
\gls{ea} and \gls{nsga} the difference is not statistically significant which is
interesting when compared to the projections in Figure~\ref{fig:map_evolution}.
This shows that while \gls{nsga} is able to find more diverse solutions their
performance are not enough to offset the better-performing solutions found
through the single objective \gls{ea} on the \gls{qd}-score metric. To elucidate
this difference further, we have plotted cumulative coverage which shows the
number of filled morphological niches, normalized to the maximum number of
possible niches, in Figure~\ref{fig:coverage}. This figure, together with maximum
fitness, shows the trade-off between finding diverse solutions, \gls{nsga}, and finding high
performing solutions, \gls{ea}.

\begin{figure}
	\includegraphics{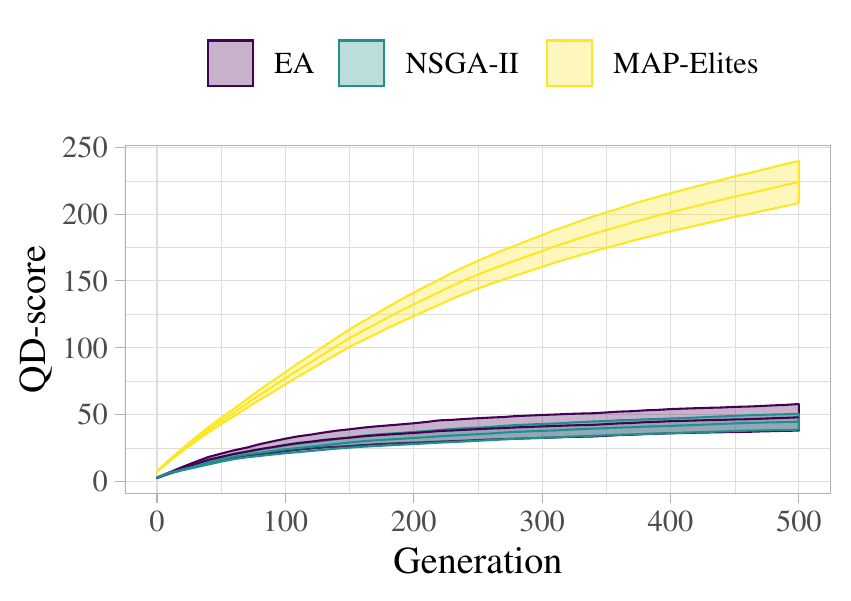}
	\vspace*{-20pt}
	\caption{\Gls{qd}-score over generational time. The average score is
	shown together with a $95\%$ confidence interval. Pairwise Wilcoxon Rank
	sum testing shows significant differences between \gls{map-elites} and
	the two other search algorithms with $p < 2e^{-16}$ in the final
	generation.}
	\label{fig:qd_score}
\end{figure}

\begin{figure}
	\includegraphics{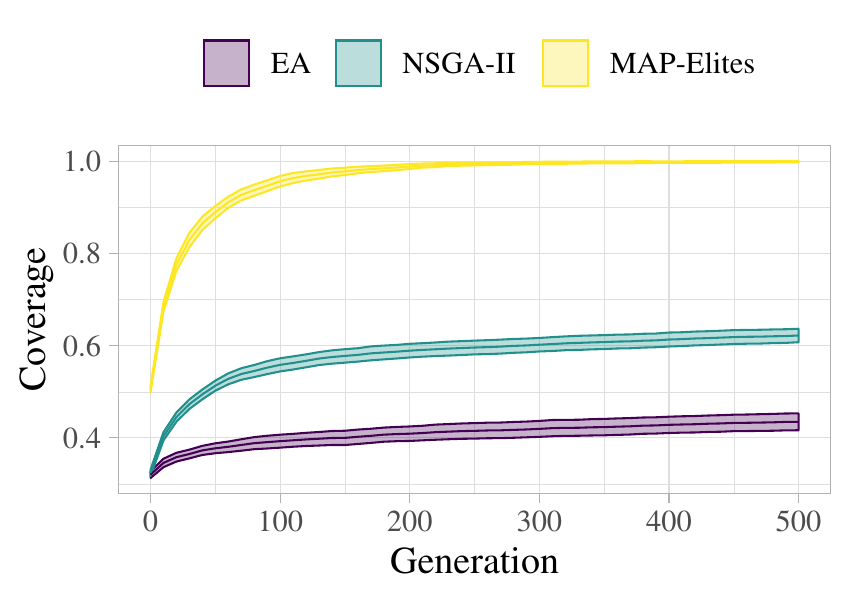}
	\vspace*{-20pt}
	\caption{Cumulative coverage, the number of filled cells normalized to
	the maximum possible filled, throughout evolution. The mean and a $95\%$
	confidence interval is shown. The difference between all search
	algorithms in the last generation is statistically significant according
	to a Wilcoxon Rank Sum test using Holm correction.}
	\label{fig:coverage}
\end{figure}

Although Figure~\ref{fig:map_evolution} gives an overview of how the different
search algorithms unfold, it is not well suited to show the distribution of the
morphologies in the population over time. In Figure~\ref{fig:histogram} the full
distribution of morphologies are shown. The figure is comprised of individual
vertical bars that show the number of each type of module, where each bar is
normalized to sum to one. The figure shows that \gls{map-elites} and \gls{nsga}
are able to evolve diverse solutions in most runs, while the single objective
\gls{ea} focuses on fewer morphologies. Furthermore it can be seen that
\gls{nsga} has more fluctuation in the population over time compared to
\gls{map-elites}. Lastly, it is interesting to note that the single objective
\gls{ea} is able to rediscover morphologies, which can be seen as bands of
colors that appear, vanish and then re-appear throughout the search. 

\begin{figure*}
	\begin{subfigure}[t]{0.30\textwidth}
		\includegraphics{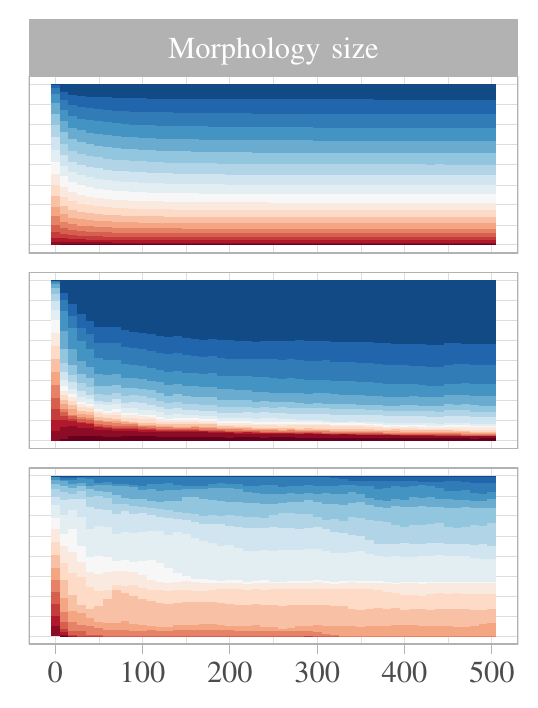}
		\label{fig:hist_all}
	\end{subfigure}
	\hspace{-8pt}
	\begin{subfigure}[t]{0.65\textwidth}
		\includegraphics{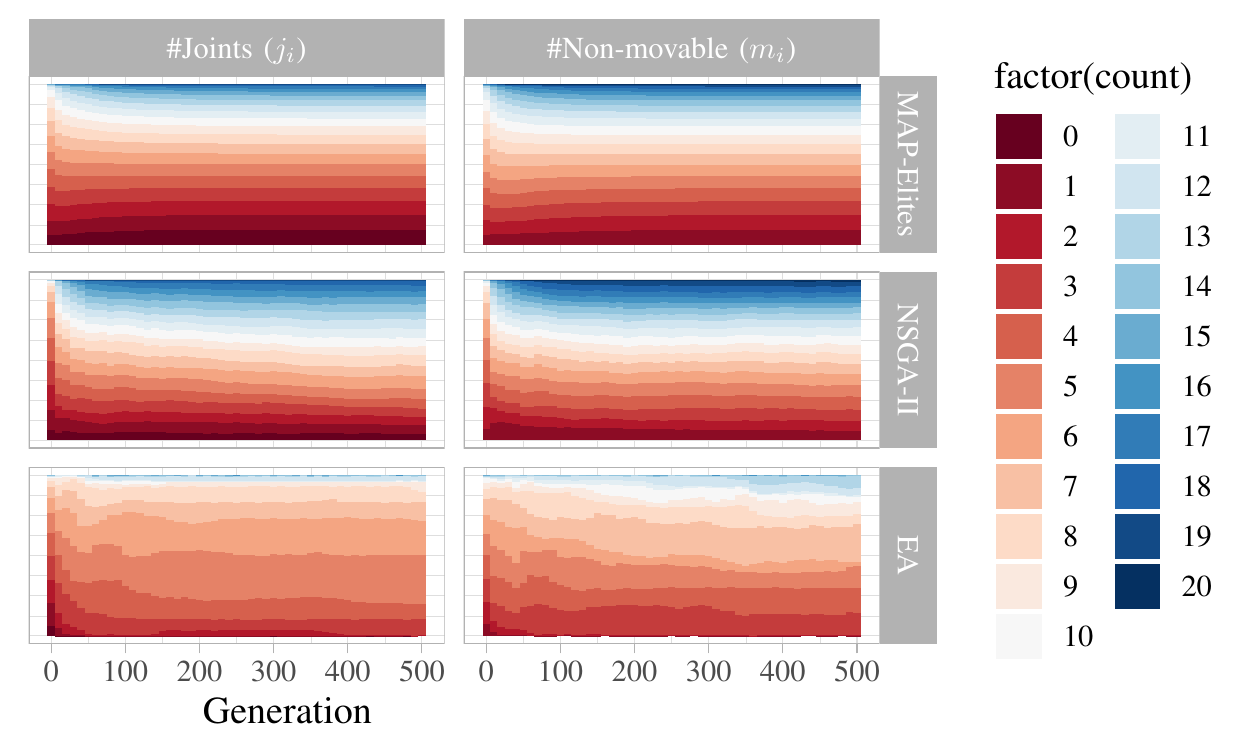}
		\label{fig:hist_split}
	\end{subfigure}
	\vspace*{-20pt}
	\caption{Distribution of modules in the population over time separated
	by search algorithm. On the left the full size of the modular robots is
	displayed while the right shows the size separated into the two
	different module types. The figure is a sequential collection of
	vertical histogram bars that show the normalized distribution of modules
	at intervals during the search. The underlying data is collected from
	all $30$ runs.}
	\label{fig:histogram}
\end{figure*}

\section{Discussion}
The results for the locomotion task in this paper show that both the single
objective \gls{ea} and \gls{map-elites} are capable of promoting the same
quality. While \gls{nsga} did not achieve the same fitness, the difference
was not immense and can presumptively be attributed to the number of dimensions
to optimize and limitations of the initial parameter sweep. This shows that for
simple locomotion, all three search algorithms can be useful tools in modular
\gls{er}.

When looking at the evolution of fitness over time, illustrated in
Figure~\ref{fig:fitness}, we can see that fitness of the single objective
\gls{ea} quickly grows in contrast with the two other algorithms. This is in
line with previous results comparing \glspl{ea} and \gls{qd} algorithms where
\gls{qd} algorithms tend to have slower growth~\cite{gaier2017data}. In the
results shown here the growth is likely due to the relatively easier task of
finding new morphological niches to occupy compared to increasing fitness of
already discovered solutions. If we take the number of filled niches, shown in
Figure~\ref{fig:coverage}, into account we can see the rapid increase in filling
out new niches taking place early in the search for \gls{map-elites} further
reinforcing this explanation. This could indicate that the \gls{map-elites}
search could benefit from adding \textit{curiosity}~\cite{cully2017quality} or
\textit{dynamic mutation}~\cite{nordmoen2018dynamic}. From
Figure~\ref{fig:fitness} it can also be seen that \gls{map-elites} has a lower
between-run variance than the two other algorithms which indicates that
\gls{map-elites} more consistently find high performing solutions.

In addition to the focus on fitness, it can be beneficial to encourage diversity
to avoid premature convergence and increase robustness to noise.
Figure~\ref{fig:map_evolution}, Figure~\ref{fig:qd_score}, and
Figure~\ref{fig:coverage} explored how the three search algorithms compared when
the populations are projected into a grid of the two morphological descriptors
and how these projections can be summarized. The projections show that
\gls{map-elites} is better at exploring the search landscape of different
morphologies, finding nearly every possible morphology in all repetitions of the
search. While \gls{nsga} is able to find every expression across all runs, as
can be seen in the middle row of Figure~\ref{fig:map_evolution}, it did not
manage the same reliability per run as \gls{map-elites}. This is most likely
caused by the complex Pareto front which \gls{nsga} maintains in its population,
which works well for optimization problems, but is more difficult to apply to
explicit morphological diversity measures. It is also possible that \gls{nsga}'s
diversity maintenance conflicts with morphological diversity resulting in a
slower search.

To understand how the population evolves in each search algorithm we plotted the
distribution of morphologies in Figure~\ref{fig:histogram}. From this plot, it
can be seen that both \gls{map-elites} and \gls{nsga} quickly fills out all
morphological niches. The same trend was seen in Figure~\ref{fig:coverage} where
both quickly converge to their respective maximum in contrast to fitness,
Figure~\ref{fig:fitness}, which slowly grows throughout the experiment. In
comparison we can see that \gls{nsga} has more variation across morphologies and
the focus of the search is on larger morphologies, both in terms of joints and
non-movable modules. As one would expect, the single objective \gls{ea} focuses
on fewer morphologies over evolutionary time with some combination vanishing and
re-appearing. The reduction in diversity for the single objective \gls{ea} could
be a sign of premature convergence. In contrast, both \gls{map-elites} and
\gls{nsga} do not converge to a few solutions which could be a sign of a more
sound search.

The next step for this research is to more closely look into why
\gls{map-elites} is able to evolve both high quality and diverse solutions.
Early indications point to the idea of \textit{stepping-stones} where
\gls{map-elites}, due to the elitist definition of the search, is capable of
retaining and promoting better solutions that further the
search~\cite{gaier2019quality}. By understanding the genealogy of the search we
hope to generate objective statistics that can discern such details and give a
better understanding of \gls{map-elites}.

To extend on the foundation of modular robotics, experiments including \gls{qd}
algorithms and indirect encodings would be a logical extension of this work.
Additionally, testing different control schemes and morphological - behavioral
metrics could improve the results even further. 

\section{Conclusion}
In this paper we compared a single objective \gls{ea} with two diversity
promoting search algorithms, a \gls{moea} and a \gls{qd} algorithm, on their
capacity for evolving a diverse set of high performing solutions over time on
the difficult task of optimizing morphology and control in modular robotics.
The result shows that the different algorithms have nearly the same capacity for
quality, however, morphological diversity can be greatly improved, without
affecting the maximum fitness obtained, by utilizing morphological descriptors
to aid the search. The results also show that the method of applying
morphological descriptors can impact performance and \gls{map-elites}, due to
its simplicity, is well suited for application in modular robotics, achieving
both high fitness and large diversity.

The work in this paper is a supplement both to the work on modular robotics and
application areas for \gls{qd} algorithms. By demonstrating that both high
fitness as well as large diversity can be promoted simultaneously, future
research on modular robotics can evolve repertoires of morphologies that can be
exploited for different purposes. For \gls{qd} this work opens up a new
application domain in which rapid exploration of real-world robotics can be
experimented with.

\section*{Acknowledgements}
The experiments were performed on resources provided by UNINETT Sigma2 - the
National Infrastructure for High Performance Computing and Data Storage in
Norway. This work was partially supported by the Research Council of Norway
through its Centres of Excellence scheme, project number 262762.

\footnotesize
\bibliographystyle{IEEEtran}
\bibliography{modular}
\end{document}